# Exploring the Effects of Population and Employment Characteristics on Truck Flows: An Analysis of NextGen NHTS Origin-Destination Data[†]

**Majbah Uddin, PhD, EIT,[1] Yuandong Liu, PhD,[2] and Hyeonsup Lim, PhD[3]**

[1]National Transportation Research Center, Oak Ridge National Laboratory, 1 Bethel Valley Rd, Oak Ridge, TN 37830; e-mail: uddinm@ornl.gov
[2]National Transportation Research Center, Oak Ridge National Laboratory, 1 Bethel Valley Rd, Oak Ridge, TN 37830; e-mail: liuy@ornl.gov
[3]National Transportation Research Center, Oak Ridge National Laboratory, 1 Bethel Valley Rd, Oak Ridge, TN 37830; e-mail: limh@ornl.gov

**ABSTRACT**

Truck transportation remains the dominant mode of US freight transportation because of its advantages, such as the flexibility of accessing pickup and drop-off points and faster delivery. Because of the massive freight volume transported by trucks, understanding the effects of population and employment characteristics on truck flows is critical for better transportation planning and investment decisions. The US Federal Highway Administration published a truck travel origin-destination data set as part of the Next Generation National Household Travel Survey program. This data set contains the total number of truck trips in 2020 within and between 583 predefined zones encompassing metropolitan and nonmetropolitan statistical areas within each state and Washington, DC. In this study, origin-destination–level truck trip flow data was augmented to include zone-level population and employment characteristics from the US Census Bureau. Census population and County Business Patterns data were included. The final data set was used to train a machine learning algorithm-based model, Extreme Gradient Boosting (XGBoost), where the target variable is the number of total truck trips. Shapley Additive ExPlanation (SHAP) was adopted to explain the model results. Results showed that the distance between the zones was the most important variable and had a nonlinear relationship with truck flows.

[†]This manuscript has been authored by UT-Battelle, LLC, under contract DE-AC05-00OR22725 with the US Department of Energy (DOE). The US government retains and the publisher, by accepting the article for publication, acknowledges that the US government retains a nonexclusive, paid-up, irrevocable, worldwide license to publish or reproduce the published form of this manuscript, or allow others to do so, for US government purposes. DOE will provide public access to these results of federally sponsored research in accordance with the DOE Public Access Plan (http://energy.gov/downloads/doe-public-access-plan).

# INTRODUCTION

Trucks, which are the dominant mode of US freight transportation, play a critical role in the movement of goods in the United States. Truck transportation has the flexibility of accessing pickup and drop-off points and faster delivery times. According to Freight Analysis Framework (US Department of Transportation 2022), more than 64.8% of freight in tonnage (carrying over 12 billion tons) was transported by trucks in 2020, and that share of truck transportation is projected to be even higher in forecasted years (66.8% in 2050). Additionally, trucks also contributed to 597 million tons of multimodal freight movements (e.g., truck–rail and truck–water), accounting for multiple modes involving trucks explicitly based on the *2017 Commodity Flow Survey* data (US Census Bureau 2020). With the major role of trucks to transport massive freight volumes and connect different transportation modes and networks, understanding truck movements is critical for better transportation planning and investment decisions at different geographical levels. Truck movements also have a major effect on the economy, environment, highway congestion, and transportation safety.

The existing literature on freight/truck flows can be summarized into two broad categories. The first category is the estimation of the truck trip generation model at regional scales. The second category is truck trip rates at a disaggregated level, such as for different types of land use, industries, and special facilities (Al-Deek et al. 2020; Shin and Kawamura 2006; Holguin-Veras 2022; McCormack 2010). Given the scope of this paper, the studies related to the estimation of trip generation at regional scales were reviewed. In particular, the variables and attributes that affect freight and truck trip generation at the regional scale were identified and summarized. Employment and related attributes are some of the most important factors affecting regional trip generation (Batida and Holguín-Veras 2009, Doustmohammadi et al. 2019; Kulpa 2014; Motuba and Tolliver 2017; Sánchez-Díaz and Holguín-Veras 2016). Batida and Holguín-Veras (2009) found that the number of employees, industry segment, commodity type, facility type, and total sales were statistically significant in estimating freight generation. Sánchez-Díaz and Holguín-Veras (2016) investigated the spatial effects of freight trip attraction. Their study indicated that the employment in different sectors was an important predictor. They found that regardless of having the lowest employment, retail establishments tend to have the largest freight attractions. In addition to employment in different sectors, Doustmohammadi et al. (2019) used truck Global Positioning System (GPS) data and developed a community-specific truck trip generation model. Other variables that were found to have effect on truck and freight trip generations include land use (Brogan 1979; Lawson et al. 2012), number of inhabitants, number of truck parking, and region types (i.e., urban or rural) (Kulpa 2014).

To the best of the authors' knowledge, this study is the first to explore the effects of population and employment characteristics on truck flows based on the *2020 Next Generation National Household Travel Survey* (NextGen NHTS) origin-destination (OD) data. The truck flows were available at zone-to-zone OD levels. Population data were obtained from the US Census Bureau, and employment characteristics (i.e., number of establishments, number of employees, and annual payroll) were obtained from the US Census Bureau's County Business Patterns (CBP) data set. An Extreme Gradient Boosting (XGBoost) algorithm, one of the most popular algorithms for regression problems, was adopted to train a predictive model. To interpret the model outcomes, Shapley Additive ExPlanation (SHAP) value-based analyses were conducted. These analyses provided indications of how much each explanatory variable contributed to truck flows, either positively or negatively.

**DATA SOURCES**

In this study, three data sources were used: NextGen NHTS OD data (Federal Highway Administration [FHWA] 2022), CBP data (US Census Bureau 2022a), and census population estimates at the county level (US Census Bureau 2022b). A brief description of the first two data sources is provided in this section.

**NextGen NHTS OD Data**

FHWA recently published 2020 OD data products as part of NextGen NHTS. The OD trips were based on passively collected mobility data using in-vehicle and smartphone applications. The 2020 OD data products contained annualized trip counts for truck and passenger travel within and between 583 FHWA zones. These zones consisted of 447 state-specific metropolitan statistical areas and 136 new zones created from the remaining nonmetropolitan statistical areas. This study used the national truck OD data. The truck trips were estimated using truck GPS data and validated using other data sources (e.g., American Transportation Research Institute and INRIX). The trips reflected movements by freight trucks and light-duty transportation trucks used for both intercity and local deliveries. These trucks do not include pickup trucks. Figure 1 shows the top 1,000 OD pairs for trucks in 2020, excluding intrazonal flows.

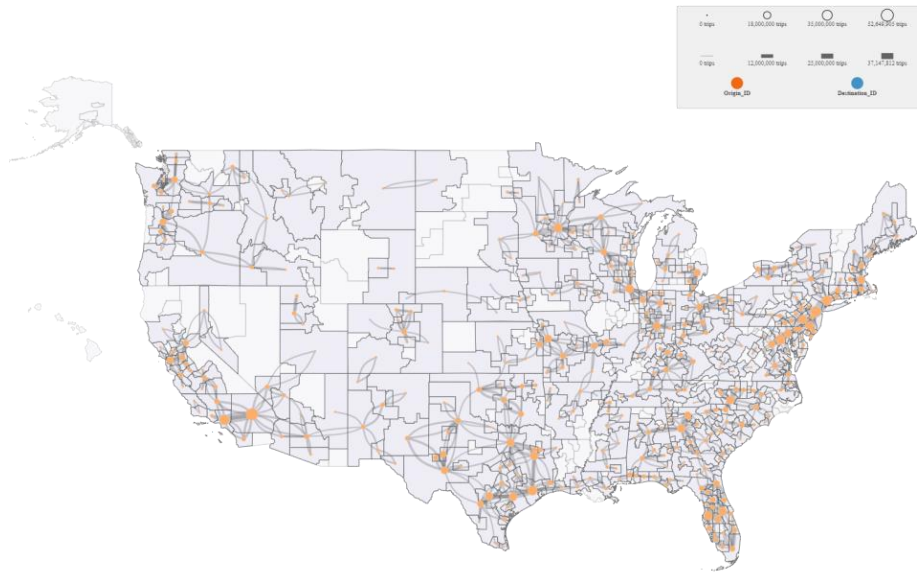

**Figure 1. Top 1,000 OD pairs for trucks in 2020, excluding intrazonal flows (FHWA 2022).**

The truck data set contained more than 339,000 records. Two filtering steps were applied. First, records where flow origins and destinations were in Alaska and Hawaii were deleted. Second, only nonzero total truck trip records were kept. The final data set had 209,851 data records. Only total truck trips and associated FHWA origin and destination zones were considered in this study. The data set was augmented with additional features. For example, the great circle distances (GCDs) between origin and destination zones were calculated using zone centroids. Table 1 presents descriptive statistics of the variables considered.

**CBP**

CBP is a US Census Bureau program that provides annual subnational economic data by industry. CBP covers over 6,000,000 single-unit and 1,800,000 multiunit establishments. An establishment is defined as a single physical location at which business is conducted or industrial or services are performed. A company could consist of one or more establishments. The CBP data series used in this study included the number of establishments, number of employees during the week of March 12, and annual payroll values. County-level data were aggregated into FHWA zones so that these data could be merged with the truck OD data.

**METHODOLOGY**

**Regression of Truck Flows Using XGBoost**

The XGBoost algorithm was used to understand annual total truck flows. XGBoost can construct boosted trees efficiently. It also builds a single learner by combining weak learners interactively. In this study, Python's XGBoost library was used to train and evaluate the model (XGBoost 2022).

**Interpreting Regression of Truck Flows Using SHAP**

To interpret the developed XGBoost model, SHAP was used. SHAP, which is based on Shapley values, summarized the contribution of each feature to the predicted truck flows. In this study, Python's SHAP library was used to interpret the XGBoost model (SHAP 2022).

**Implementation**

A model was first trained on the training data set using an off-the-shelf XGBoost algorithm and then validated on the test data set. A grid search approach was used to tune hyperparameters that yielded a reasonable model performance. Once a proper predicted model was trained, the model was applied to the training data set. Lastly, SHAP values for each feature on each individual prediction were calculated. Summary plots were generated to present the relation between feature importance and feature effects. Additionally, dependence plots were generated to present the effect of a single feature on the predicted values.

**Table 1. Descriptive statistics for the variables considered**

| Variable | Description | Mean | Median | Min | Max |
|---|---|---:|---:|---:|---:|
| **Target variable** | | | | | |
| annual_total_trips | Annual truck trip total estimates | 83,568.7 | 278 | 30 | 466,407,788 |
| log_truck_trips | log of annual truck trip total estimates | 6.1 | 5.6 | 3.4 | 20.0 |
| **Explanatory variables** | | | | | |
| origin_zone | Unique identifier for the origin zone (583 zones) | — | — | — | — |
| destination_zone | Unique identifier for the destination zone (583 zones) | — | — | — | — |
| GCD | GCD between origin and destination zone centroid (miles) | 760.7 | 660.3 | 3.5 | 2,785.4 |
| orig_pop | Population total for the origin zone | 710,678.5 | 293,927 | 5,168 | 13,827,145 |
| dest_pop | Population total for the destination zone | 726,716.9 | 295,189 | 5,168 | 13,827,145 |
| orig_est | Total number of establishments for the origin zone | 17,041.9 | 6,550 | 83 | 395,339 |
| log_orig_est | log of total number of establishments for the origin zone | 8.9 | 8.8 | 4.4 | 12.9 |
| dest_est | Total number of establishments for the destination zone | 17,450.9 | 6,702 | 83 | 395,339 |
| log_dest_est | log of total number of establishments for the destination zone | 8.9 | 8.8 | 4.4 | 12.9 |
| orig_emp | Total employees for the origin zone | 277,258.9 | 97,227 | 895 | 5,963,872 |
| log_orig_emp | log of total employees for the origin zone | 11.6 | 11.5 | 6.8 | 15.6 |
| dest_emp | Total employees for the destination zone | 284,287.6 | 99,519 | 895 | 5,963,872 |
| log_dest_emp | log of total employees for the destination zone | 11.7 | 11.5 | 6.8 | 15.6 |
| orig_ap | Total annual payroll for the origin zone (×$1,000) | 15,610,480.7 | 4,222,250 | 37,441 | 462,483,145 |
| log_orig_ap | log of total annual payroll for the origin zone | 15.4 | 15.3 | 10.5 | 20.0 |
| dest_ap | Total annual payroll for the destination zone (×$1,000) | 16,076,842.8 | 4,314,267 | 37,441 | 462,483,145 |
| log_dest_ap | log of total annual payroll for the destination zone | 15.4 | 15.3 | 10.5 | 20.0 |

## RESULTS AND DISCUSSION

### XGBOOST Modeling

The model was trained on 70% of the randomly selected data, and the remaining 30% were used as a testing data set. A 10-fold cross-validation was performed to evaluate the stability of the model performance. The grid search yielded the following hyperparameters: maximum depth of a tree, *max_depth* = 10; minimum sum of instance weight needed in a child, *min_child_weight* = 6; step size, *eta* = 0.01; subsample ratio of the training instances, *subsample* = 0.8; and subsample ratio of columns when constructing each tree, *colsample_bytree* = 1.0. Notably, a number of features had skewed data distribution. For that reason, log-transformation was done (refer to Table 1 for descriptive statistics of these transformed features). The trained model had a root mean squared log error value of 0.71 and an R-squared value of 0.80. These metrics indicated a good model fit.

### Model Interpretation Using SHAP

For model interpretation based on SHAP values, the global feature importance of each feature was explored first. Figure 2 lists the most significant features in descending order. The top features contributed more to the model than the bottom features and thus had higher predictive power. As shown in the figure, GCD was the most important feature. Additionally, the figure highlights the correlations of features with the target. The red color means the feature was positively correlated with the target, and the blue color means the feature was negatively correlated with the target. GCD, the log of annual payroll in the origin zone, and the log of annual payroll in the destination zone were found to have a negative correlation with truck flows. Conversely, populations at the origin and destination, origin zone, destination zone, the log of the number of employees in the destination zone, the log of the number of employees in the origin zone, and the log of the number of establishments in the origin and destination zones were found to have a positive correlation with truck flows.

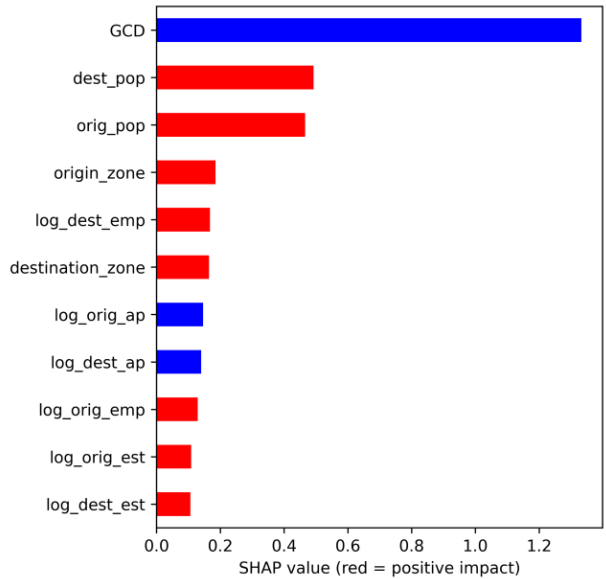

**Figure 2. Global feature importance.**

The SHAP values in Figure 3 illustrate the contribution of each feature to every individual predicted value for the target. A higher or positive SHAP value indicates a more influential contribution to the truck flow predicted value, and a lower or negative SHAP value indicates less influential contributions to the truck flow predicted value. Moreover, the blue to red color scheme represents the feature value from low to high, respectively. As the value of GCD decreased, the predicted value of truck flows increased. Also, a low GCD had a stronger positive effect on truck flows than a high GCD with a negative effect on truck flows. With a higher population in a destination zone, truck flows were higher. A high population in a destination zone had a stronger positive effect on truck flows than that of a low population in a destination zone with a negative effect on truck flows. The population in an origin zone exhibited a similar pattern. However, population in origin zone had a stronger effect on truck flows compared with population in a destination zone. With a higher number of employees in a destination zone, truck flows increased, but the opposite occurred when the number of employees was lower. Number of employees in an origin zone exhibits a similar pattern. However, a low number of employees in an origin zone had a stronger negative effect on truck flows than that of a high number of employees with a positive effect. Annual payrolls in both origin and destination zones showed a similar pattern in which higher payroll led to lower truck flows. Lastly, the number of establishments in both origin and destination zones suggest that higher numbers of establishments led to higher truck flows.

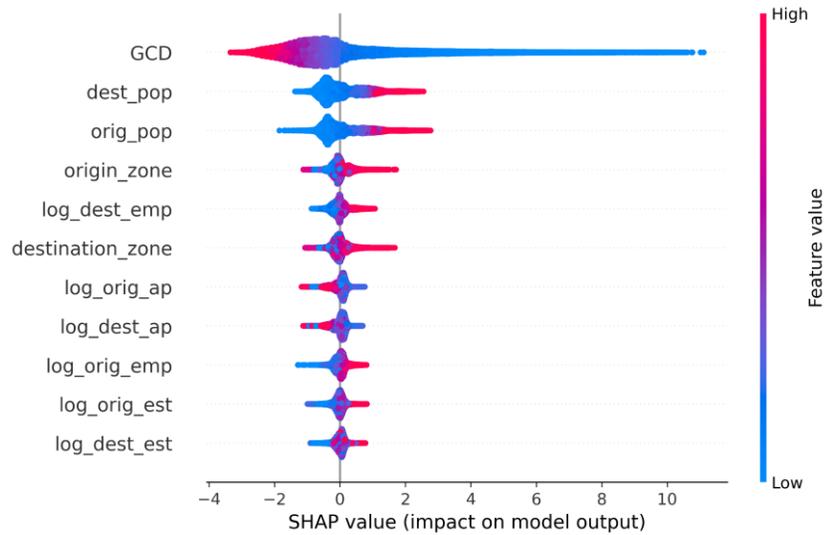

**Figure 3. Local explanation summary.**

Figure 4 shows the partial dependence plots for all the features included in the model. These plots illustrate the features' marginal effects on the model output. Notably, a positive SHAP value indicates higher truck flows, and a negative SHAP value indicates lower truck flows. Figure 4(a) shows that GCD had a nonlinear relationship with truck flows. When GCD was approximately 500 miles or smaller, the SHAP value was positive, but the opposite happened when GCD was larger than 500 miles. This suggests that when GCD was below a certain threshold (i.e., 500 miles), truck flows were higher, whereas truck flows were lower when GCD was above the threshold. Population in both origin and destination zones had nonlinear relationships with truck flows. When population in a destination zone was approximately 500,000 or lower and population in an origin zone was approximately 400,000 or lower, the SHAP values were negative, meaning that truck flows were lower. Truck flows were higher above these population thresholds. The number of employees in origin and destination zones and the number of establishments in origin and destination zones indicated somewhat linear relationships with truck flows. For these features, truck flows were higher above their average value and lower below the average value. For example, when the number of employees in a destination zone was approximately above 120,570, truck flows were higher. Annual payrolls in origin and destination zones showed a similar relationship with truck flows. The SHAP values were positive when annual payrolls were approximately between $4,400,000 and $8,900,000,000. SHAP values were negative when payrolls were outside of these payroll ranges.

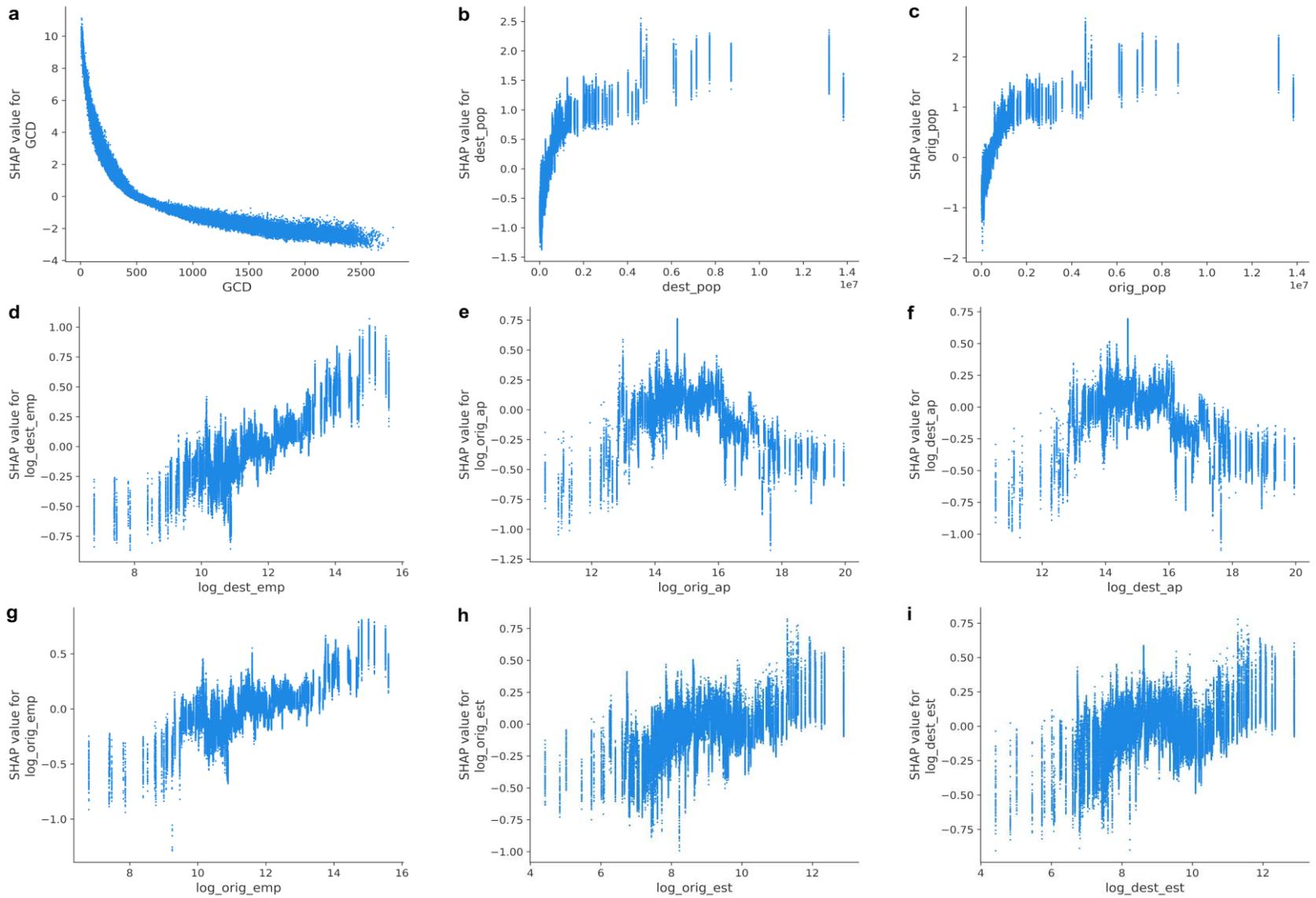

**Figure 4. Partial dependence plot: (a) GCD, (b) dest_pop, (c) orig_pop, (d) log_dest_emp, (e) log_orig_ap, (f) log_dest_ap, (g) log_orig_emp, (h) log_orig_est, and (i) log_dest_est.**

The interaction effects between features were also explored. Figure 5 shows only a few interesting interaction effects. Figure 5(a) plots the SHAP interaction values between GCD and the log of the number of establishments in an origin zone. When GCD was above 500 miles, on average, a higher number of establishments in an origin zone had a strong negative association to truck flows compared with a lower number of establishments' association to truck flows. However, when GCD was below 500 miles, the effect of the number of establishments in an origin zone was somewhat the same (i.e., higher or lower number of establishments both were positively associated with truck flows).

Figure 5(b) plots the SHAP interaction values between population and annual payroll in a destination zone. The SHAP value increased significantly to positive values when the population in a destination zone was between 0 and 500,000. These values were associated with lower annual payroll in a destination zone. For a population higher than 500,000, the SHAP value increased at a flat rate, which was associated with higher annual payroll in a destination zone. Figure 5(c) plots SHAP interaction values between the population in an origin zone and annual payroll in a destination zone. The pattern for interaction values did not show any trend.

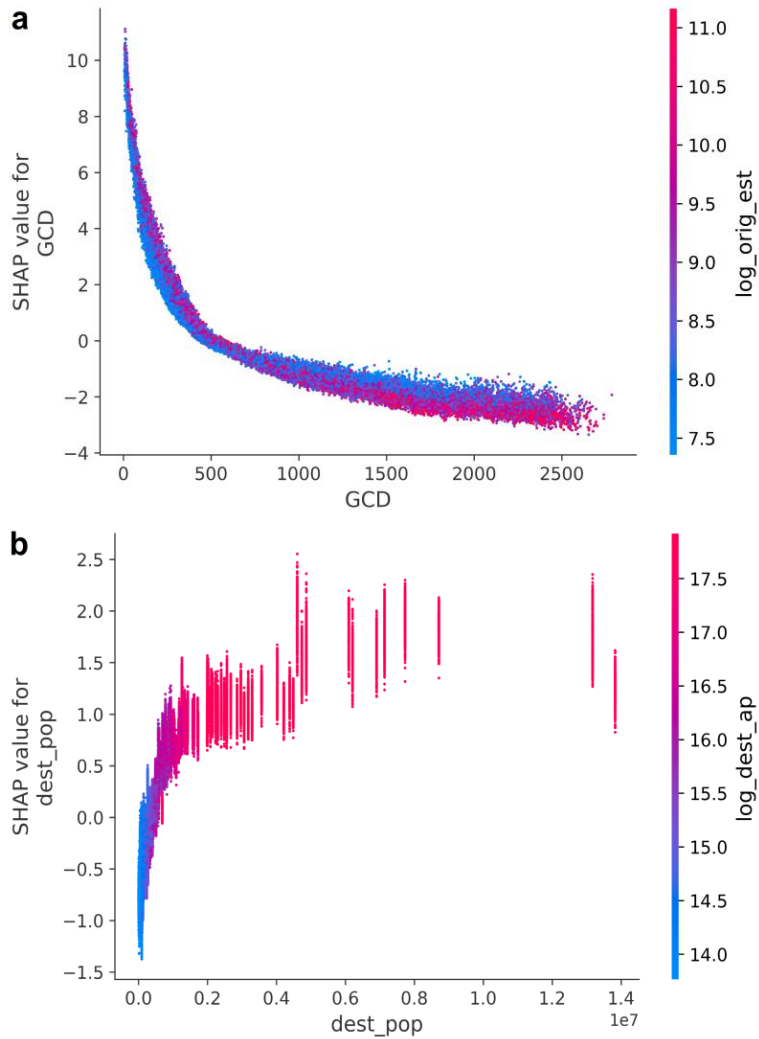

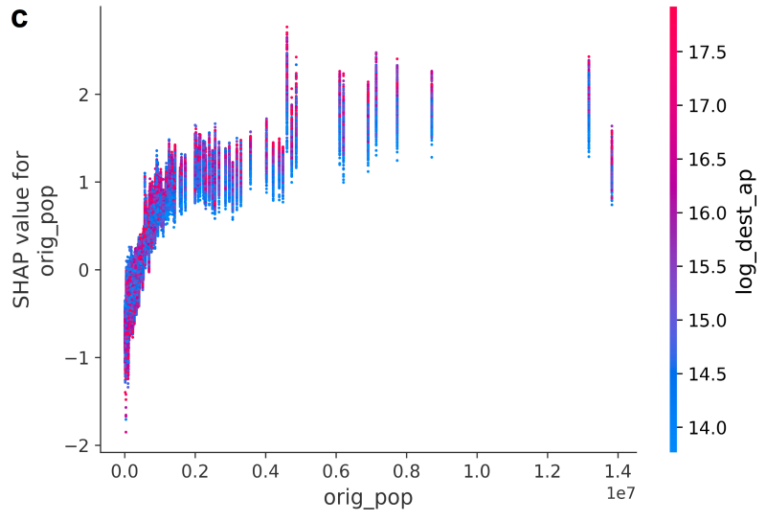

**Figure 5. Interaction effects between (a) GCD and log_orig_est, (b) dest_pop and log_dest_ap, and (c) orig_pop and log_dest_ap.**

## CONCLUSION

This study explored the effects of population and employment characteristics on truck flows. The OD-level truck flows (i.e., trips) were obtained from the NextGen NHTS OD truck data set. These truck flows were provided at 583 zones (i.e., regions). Population data were obtained from county-level census estimates. Employment characteristics (number of establishments, number of employees, and annual payroll) were obtained from the CBP data set. A master data set was constructed using these three sources, and an interpretable machine learning framework was employed to explain the outcomes from the model. Particularly, a predictive model was trained based on the XGBoost algorithm, and then a SHAP framework was implemented to interpret the model.

Based on the feature importance, a few of the top important features were GCD, populations in origin and destination zones, and the number of employees in a destination zone. SHAP partial dependence plots were analyzed to explore the linear and nonlinear relationships between features and target (truck flows). Additionally, interaction effects were analyzed in a few features where interesting findings were observed. Notably, when GCD was below 500 miles, truck flows were higher. Truck flows were lower when GCD was above 500 miles. Also, GCD interacted mostly with the number of establishments in an origin zone. In future works, the authors would like to extend the framework to better model interaction and/or nonlinear effects in the features that were considered in the study.

Bastida, C., and Holguín-Veras, J. (2009). "Freight generation models: Comparative analysis of regression models and multiple classification analysis." *Transp. Res. Rec.*, 2097(1), 51–61.

Brogan, J. D. (1979). "Development of truck trip-generation rates by generalized-land use categories." *Transp. Res. Rec.*, 716, 38–43.

US Census Bureau. (2020). "2017 Commodity Flow Survey." <https://www.census.gov/data/tables/2017/econ/cfs/aff-2017.html> (Nov. 14, 2022).

US Census Bureau. (2022a). "County Business Pattern." <https://www.census.gov/programs-surveys/cbp.html> (Nov. 10, 2022).

US Census Bureau (2022b). "Population Estimate." <https://www.census.gov/topics/population.html> (Nov. 10, 2022).

Doustmohammadi, M., Anderson, M., and Doustmohammadi, E. (2019). "Regression Analysis to Create New Truck Trip Generation Equations for Medium Sized Communities." *Curr. Urban Stud.*, 7(03), 480.

FHWA (US Federal Highway Administration). (2022). "2020 NextGen NHTS National Truck OD Data." <https://nhts.ornl.gov/od/> (Sep. 1, 2022).

FHWA (US Federal Highway Administration). (2022). "NextGen NHTS OD Data Portal." <https://nhts.ornl.gov/od/summary/> (Nov. 14, 2022).

Holguin-Veras, J., López-Genao, Y., and Salam, A. (2002). "Truck-trip generation at container terminals: Results from a nationwide survey." *Transp. Res. Rec.*, 1790(1), 89–96.

Kulpa, T. (2014). "Freight truck trip generation modelling at regional level." *Procedia-Social and Behavioral Sciences*, 111, 197–202.

Lawson, C. T., Holguín-Veras, J., Sánchez-Díaz, I., Jaller, M., Campbell, S., and Powers, E. L. (2012). "Estimated generation of freight trips based on land use." *Transp. Res. Rec.*, 2269(1), 65–72.

McCormack, E., Ta, C., Bassok, A., and Fishkin, E. (2010). *Truck trip generation by grocery stores* (No. TNW2010-04), United States Department of Transportation.

Motuba, D., and Tolliver, D. (2017). "Truck trip generation in small-and medium-sized urban areas." *Transp. Plann. Technol.*, 40(3), 327–339.

Sánchez-Díaz, I., Holguín-Veras, J., and Wang, X. (2016). "An exploratory analysis of spatial effects on freight trip attraction." *Transportation*, 43(1), 177–196.

SHAP (2022). "Python Package." <https://shap.readthedocs.io/en/latest/index.html> (Nov. 1, 2022).

Shin, H. S., and Kawamura, K. (2006). *Development of disaggregate-level truck trip generation model: Case study of furniture chain stores*, TRB 85th Annual Meeting Compendium of Papers CD-ROM, Washington, DC.

US Department of Transportation. (2022). "Freight Analysis Framework Version 5.4." <https://www.bts.gov/faf> (Nov. 14, 2022).

XGBoost (Extreme Gradient Boosting). (2022). "Python Package." <https://xgboost.readthedocs.io/en/stable/python/index.html> (Oct. 15, 2022).
12